\def\year{2019}\relax
\documentclass[letterpaper]{article} 
\usepackage{aaai19}  
\usepackage{times}  
\usepackage{helvet}  
\usepackage{courier}  
\usepackage{url}  
\usepackage{graphicx}  
\usepackage{multirow}
\usepackage{amsthm}
\usepackage{amsfonts}
\usepackage{MnSymbol}
\usepackage{makecell}
\usepackage{arydshln}
\usepackage{microtype}
\usepackage{amsmath}
\usepackage{epsfig}
\usepackage{subcaption}
\usepackage{textcomp}

\DeclareMathOperator*{\argmin}{arg\,min}

\newcommand{\citet}[1]{\citeauthor{#1}~\shortcite{#1}}
\newcommand{\citep}{\cite}

\usepackage{booktabs}
\usepackage{url}
\usepackage{algorithm}
\usepackage[noend]{algpseudocode}
\algdef{SE}[SUBALG]{Indent}{EndIndent}{}{\algorithmicend\ }%
\algtext*{Indent}
\algtext*{EndIndent}

\newcommand\blfootnote[1]{%
  \begingroup
  \renewcommand\thefootnote{}\footnote{#1}%
  \addtocounter{footnote}{-1}%
  \endgroup
}

\newcommand{\ourl}{Multimodal Cyclic Translation Network}
\newcommand{\ours}{MCTN}
\begin{document}
%
\title{Found in Translation: \\ Learning Robust Joint Representations by Cyclic Translations Between Modalities}

\author{Hai Pham$^1$*, Paul Pu Liang$^1$*, Thomas Manzini$^2$, Louis-Philippe Morency$^1$, Barnab\'{a}s P\'{o}czos$^1$ \\
$^1$Carnegie Mellon University, $^2$Microsoft AI \\
\texttt {\{htpham,pliang\}@cs.cmu.edu}
}

\maketitle

\begin{abstract}
Multimodal sentiment analysis is a core research area that studies speaker sentiment expressed from the language, visual, and acoustic modalities. The central challenge in multimodal learning involves inferring joint representations that can process and relate information from these modalities. However, existing work learns joint representations by requiring all modalities as input and as a result, the learned representations may be sensitive to noisy or missing modalities at test time. With the recent success of sequence to sequence (Seq2Seq) models in machine translation, there is an opportunity to explore new ways of learning joint representations that may not require all input modalities at test time. In this paper, we propose a method to learn robust joint representations by translating between modalities. Our method is based on the key insight that translation from a source to a target modality provides a method of learning joint representations using only the source modality as input. We augment modality translations with a cycle consistency loss to ensure that our joint representations retain maximal information from all modalities. Once our translation model is trained with paired multimodal data, we only need data from the source modality at test time for final sentiment prediction. This ensures that our model remains robust from perturbations or missing information in the other modalities. We train our model with a coupled translation-prediction objective and it achieves new state-of-the-art results on multimodal sentiment analysis datasets: CMU-MOSI, ICT-MMMO, and YouTube. Additional experiments show that our model learns increasingly discriminative joint representations with more input modalities while maintaining robustness to missing or perturbed modalities.
\end{abstract}

\section{Introduction}

\begin{figure}[ht]
\centering
\includegraphics[width=.8\linewidth]{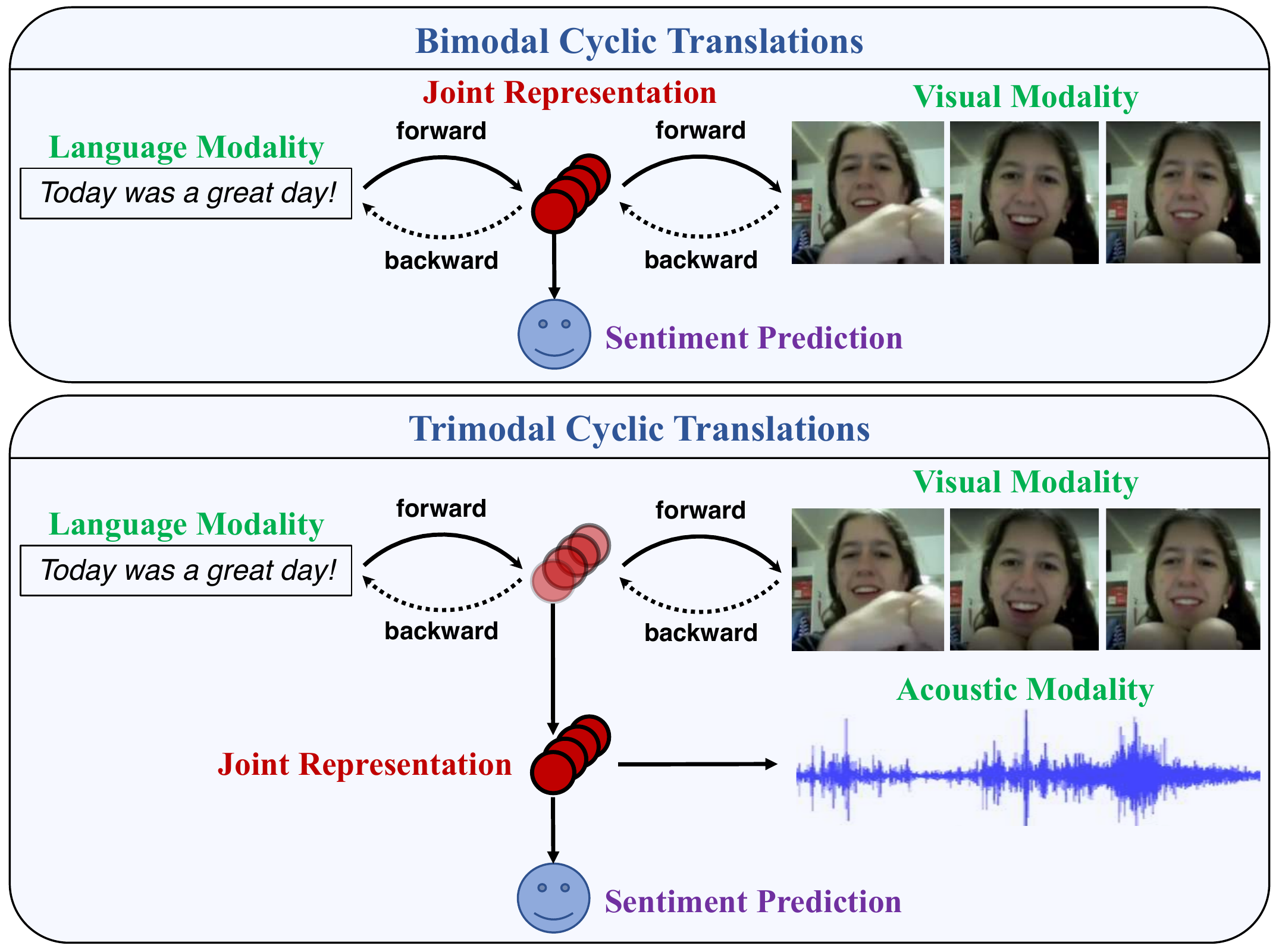}
\caption{
\small
{Learning robust joint representations via multimodal cyclic translations. Top: cyclic translations from a source modality (language) to a target modality (visual). Bottom: the representation learned between language and vision are further translated into the acoustic modality, forming the final joint representation. In both cases, the joint representation is then used for sentiment prediction.}}
\label{fig:overview}
\end{figure}

Sentiment analysis is an open research problem in machine learning and natural language processing which involves identifying a speaker's opinion~\cite{Pang:2002:TUS:1118693.1118704}. Previously, text-only sentiment analysis through words, phrases, and their compositionality can be found to be insufficient for inferring sentiment content from spoken opinions~\citep{morency2011towards}, especially in the presence of rich nonverbal behaviors which can accompany language~\citep{shaffer2018exploring}. As a result, there has been a recent push towards using machine learning methods to learn joint representations from additional information present in the visual and acoustic modalities. This research field has become known as multimodal sentiment analysis and extends the conventional text-based definition of sentiment analysis to a multimodal environment.\blfootnote{* Equal contributions} For example,~\citep{kaushik2013sentiment} explore the additional acoustic modality while~\citep{wollmer2013youtube} use the language, visual, and acoustic modalities present in monologue videos to predict sentiment. This push has been further bolstered by the advent of multimodal social media platforms, such as YouTube, Facebook, and VideoLectures which are used to express personal opinions on a worldwide scale. The abundance of multimodal data has led to the creation of multimodal datasets, such as CMU-MOSI~\citep{zadeh2016multimodal} and ICT-MMMO~\citep{wollmer2013youtube}, as well as deep multimodal models that are highly effective at learning discriminative joint multimodal representations~\citep{localglobal,factorized,chen2017msa}. Existing prior work learns joint representations using multiple modalities as input~\citep{multistage,morency2011towards,zadeh2016multimodal}. However, these joint representations also regain all modalities at test time, making them sensitive to noisy or missing modalities at test time~\citep{DBLP:conf/cvpr/Tran0ZJ17,Cai:2018:DAL:3219819.3219963}.

To address this problem, we draw inspiration from the recent success of Seq2Seq models for unsupervised representation learning~\citep{seq2seq_nn,DBLP:journals/corr/TuLSLL16}. We propose the \ourl \ model (\ours) to learn robust joint multimodal representations by translating between modalities. Figure~\ref{fig:overview} illustrates these translations between two or three modalities. Our method is based on the key insight that translation from a source modality $S$ to a target modality $T$ results in an intermediate representation that captures joint information between modalities $S$ and $T$. \ours \ extends this insight using a cyclic translation loss involving both \textit{forward translations} from source to target modalities, and \textit{backward translations} from the predicted target back to the source modality. Together, we call these \textit{multimodal cyclic translations} to ensure that the learned joint representations capture maximal information from both modalities. We also propose a hierarchical \ours \ to learn joint representations between a source modality and multiple target modalities. \ours \ is trainable end-to-end with a coupled translation-prediction loss which consists of (1) the cyclic translation loss, and (2) a prediction loss to ensure that the learned joint representations are task-specific (\textit{i.e.} multimodal sentiment analysis). Another advantage of \ours \ is that once trained with multimodal data, we \textit{only} need data from the source modality at test time to infer the joint representation and label. As a result, \ours \ is completely robust to test time perturbations or missing information on other modalities.

Even though translation and generation of videos, audios, and text are difficult~\citep{DBLP:journals/corr/abs-1710-00421}, our experiments show that the learned joint representations can help for discriminative tasks: \ours \ achieves new state-of-the-art results on multimodal sentiment analysis using the CMU-MOSI~\citep{zadeh2016multimodal}, ICT-MMMO~\citep{wollmer2013youtube}, and YouTube~\citep{morency2011towards} public datasets. Additional experiments show that \ours \ learns increasingly discriminative joint representations with more input modalities during training. 

\section{Related Work}

Early work on sentiment analysis focused primarily on written text~\citep{Pang:2002:TUS:1118693.1118704,pang2008opinion,socher2013recursive}. Recently, multimodal sentiment analysis has gained more research interest~\citep{mm_survey}.  Probably the most challenging task in multimodal sentiment analysis is learning a joint representation of multiple modalities. Earlier work used fusion approaches such as concatenation of input features ~\citep{ngiam2011multimodal,lazaridou2015combining}. Several neural network models have also been proposed to learn joint multimodal representations. \citep{multistage} presented a multistage approach to learn hierarchical multimodal representations. The Tensor Fusion Network~\citep{tensoremnlp17} and its approximate low-rank model~\citep{lowrank} presented methods based on Cartesian-products to model unimodal, bimodal and trimodal interactions. The Gated Multimodal Embedding model~\citep{chen2017msa} learns an on-off switch to filter noisy or contradictory modalities. Other models have proposed using attention~\citep{Cheng:2017:HMA:3077136.3080671} and memory mechanisms~\citep{zadeh2018memory} to learn multimodal representations.

In addition to purely supervised approaches, generative methods based on Generative Adversarial Networks (GANs)~\citep{gan} have attracted significant interest in learning joint distributions between two or more modalities~\citep{bigan,triplegan}. 
Another method for multimodal data is to develop conditional generative models~\citep{conditionalvae,variationalmultimodal} and learn to translate one modality to another. Generative-discriminative objectives have been used to learn either joint~\citep{seq2seq,kiros2014unifying} or factorized~\citep{factorized} representations. Our work takes into account the sequential dependency of modality translations and explores the effect of a cyclic translation loss on modality translations.

Finally, there has been some progress on accounting for noisy or missing modalities at test time. One general approach is to infer the missing modalities by modeling the probabilistic relationships among different modalities.~\citet{JMLR:v15:srivastava14b} proposed using Deep Boltzmann Machines to jointly model the probability distribution over multimodal data. Sampling from the conditional distributions over each modality allows for test-time inference in the presence of missing modalities.~\citet{NIPS2014_5279} trained Restricted Boltzmann Machines to minimize the variation of information between modality-specific latent variables. Recently, neural models such as cascaded residual autoencoders~\citep{DBLP:conf/cvpr/Tran0ZJ17}, deep adversarial learning~\citep{Cai:2018:DAL:3219819.3219963}, or multiple kernel learning~\citep{10.1007/978-3-642-15549-9_49} have also been proposed for these tasks. It was also found that training with modalities dropped at random can improve the robustness of joint representations~\citep{ngiam2011multimodal}. These methods approximately infer the missing modalities before prediction~\citep{Q14-1023,AAAI1714811}, leading to possible error compounding. On the other hand, \ours \ remains fully robust to missing or perturbed modalities during testing. 

\section{Proposed Approach}\label{sec:PROPAPR}

In this section, we describe our approach for learning joint multimodal representations through modality translations.

\subsection{Problem Formulation and Notation}

A multimodal dataset consists of $N$ labeled video segments defined as $\mathbf{X} = (\mathbf{X}^{l}, \mathbf{X}^{v}, \mathbf{X}^{a})$ for the language, visual, and acoustic modalities respectively. The dataset is indexed by $N$ such that $\mathbf{X} = (\mathbf{X}_1, \mathbf{X}_2, ..., \mathbf{X}_N)$ where $\mathbf{X}_i = ({\mathbf{X}_i^{l}}, {\mathbf{X}_i^v}, {\mathbf{X}_i^a}), \, 1 \leq i \leq N$. The corresponding labels for these $N$ segments are denoted as $\mathbf{y}=(y_1, y_2, ..., y_N), \, y_i \in \mathbb{R}$. 
Following prior work, the multimodal data is 
synchronized by aligning the input based on the boundaries of each word and zero-padding each example to obtain time-series data of the same length~\citep{multistage}. The $i$th 
sample
is given by ${\mathbf{X}_i^l} = ({\mathbf{w}_i}^{(1)}, {\mathbf{w}_i}^{(2)}, ..., {\mathbf{w}_i}^{(L)})$ where ${\mathbf{w}_i}^{(\ell)}$ stands for the $\ell$th word and $L$ is the length of each example. To accompany the language features, we also have a sequence of visual features ${\mathbf{X}_i^v} = ({\mathbf{v}_i}^{(1)}, {\mathbf{v}_i}^{(2)}, ..., {\mathbf{v}_i}^{(L)})$ and acoustic features ${\mathbf{X}_i^a} = ({\mathbf{a}_i}^{(1)}, {\mathbf{a}_i}^{(2)}, ..., {\mathbf{a}_i}^{(L)})$.

\subsection{Learning Joint Representations}

Learning a joint representation between two modalities $\mathbf{X}^{S}$ and $\mathbf{X}^{T}$ 
is defined by 
a parametrized function $f_{\theta}$ that returns an embedding $\mathcal{E}_{ST} = f_{\theta}(\mathbf{X}^{S},\mathbf{X}^{T})$. From there, another function $g_{w}$ is learned that predicts the label given this joint representation: $\hat{\mathbf{y}} = g_w(\mathcal{E}_{ST})$.

Most work follows this framework during both training and testing~\citep{multistage,lowrank,factorized,zadeh2018memory}. During training, the parameters $\theta$ and $w$ are learned by empirical risk minimization over paired multimodal data and labels in the training set $(\mathbf{X}^{S}_{tr},\mathbf{X}^{T}_{tr},\mathbf{y}_{tr})$:
\begin{align}
{\mathcal{E}}_{ST} &= f_{\theta}(\mathbf{X}^{S}_{tr},\mathbf{X}^{T}_{tr}), \\
\hat{\mathbf{y}}_{tr} &= g_{w}({\mathcal{E}}_{ST}), \\
\theta^*, w^* &= \argmin_{\theta,w} \mathbb{E} \ [\ell_{\mathbf{y}}(\hat{\mathbf{y}}_{tr},\mathbf{y}_{tr})].
\end{align}
for a suitable choice of loss function $\ell_{\mathbf{y}}$ over the labels ($tr$ denotes training set).

During testing, paired multimodal data in the test set $(\mathbf{X}^{S}_{te},\mathbf{X}^{T}_{te})$ are used to infer the label ($te$ denotes test set):
\begin{align}
{\mathcal{E}}_{ST} &= f_{\theta^*}(\mathbf{X}^{S}_{te},\mathbf{X}^{T}_{te}), \\
\hat{\mathbf{y}}_{te} &= g_{w^*}({\mathcal{E}}_{ST}).
\end{align}

\subsection{\ourl}

\begin{figure}[tbp]
\centering
\includegraphics[width=0.5\textwidth]{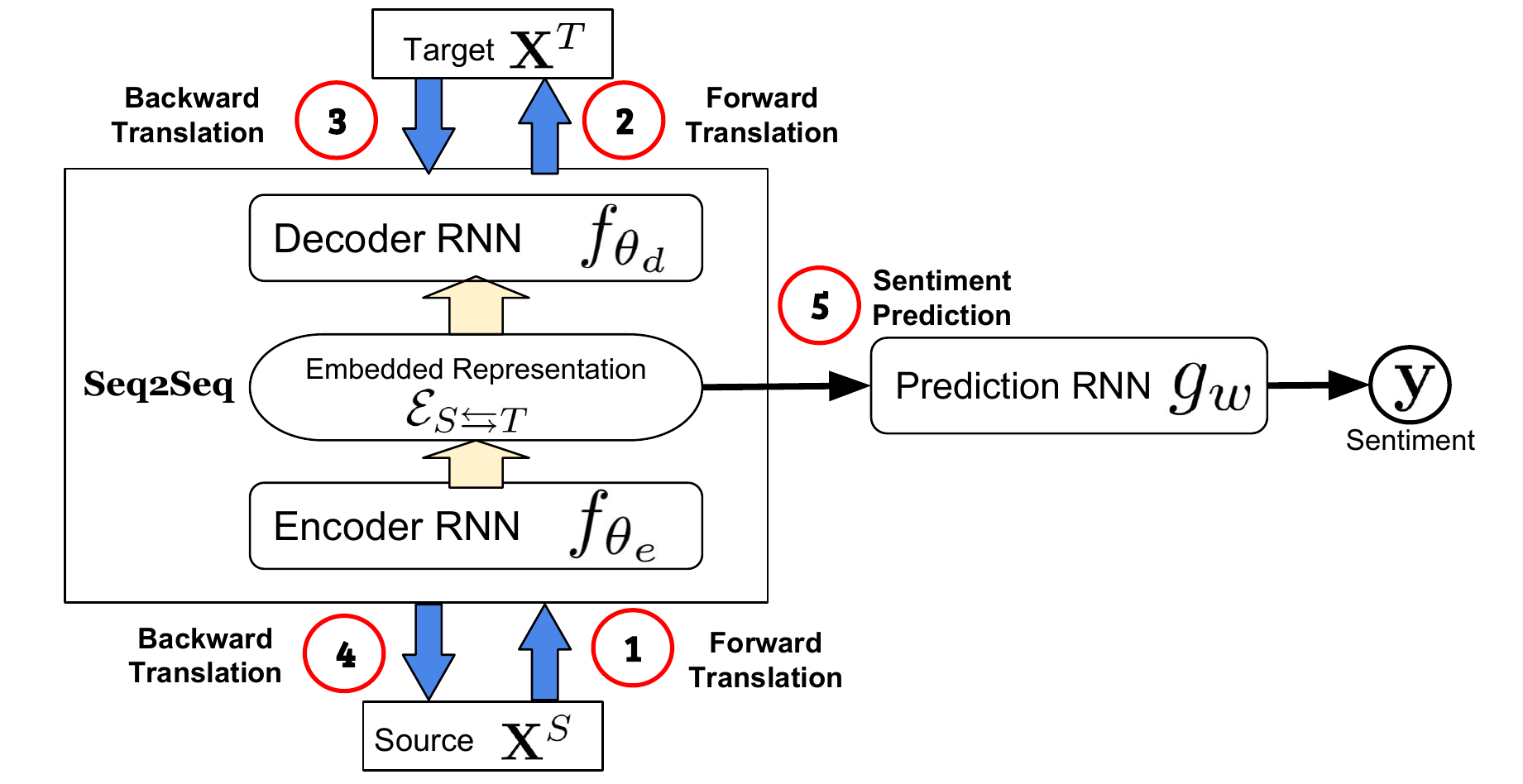}
\caption{
\small
{
\ours \ architecture for two modalities: the source modality $\mathbf{X}^{S}$ and the target modality $\mathbf{X}^{T}$. The joint representation ${\mathcal{E}}_{S \leftrightarrows T}$ is obtained via a cyclic translation between $\mathbf{X}^{S}$ and $\mathbf{X}^{T}$. Next, the joint representation ${\mathcal{E}}_{S \leftrightarrows T}$ is used for sentiment prediction. The model is trained end-to-end with a coupled translation-prediction objective. At test time, only the source modality $\mathbf{X}^{S}$ is required.}
}
\label{fig:s2s}
\end{figure}

\ourl \ (\ours) is a neural model that learns robust joint representations by modality translations. Figure~\ref{fig:s2s} shows a detailed description of \ours 
\ for two modalities. Our method is based on the key insight that translation from a source modality $\mathbf{X}^{S}$ to a target modality $\mathbf{X}^{T}$ results in an intermediate representation that captures joint information between modalities $\mathbf{X}^{S}$ and $\mathbf{X}^{T}$, but using only the source modality $\mathbf{X}^{S}$ as input
during test time. 

To ensure that our model learns joint representations that retain maximal information from all modalities, we use a cycle consistency loss~\citep{DBLP:journals/corr/ZhuPIE17} during modality translation. This method can also be seen as a variant of back-translation which has been recently applied to style transfer~\citep{DBLP:journals/corr/abs-1804-09000,DBLP:journals/corr/ZhuPIE17} and unsupervised machine translation~\citep{DBLP:journals/corr/abs-1804-07755}. We use back-translation in a multimodal environment where we encourage our translation model to learn informative joint representations but with only the source modality as input. The cycle consistency loss for modality translation starts by decomposing function $f_{\theta}$ into two parts: an encoder $f_{\theta_e}$ and a decoder $f_{\theta_d}$. The encoder takes in $\mathbf{X}^{S}$ as input and returns a joint embedding $\mathcal{E}_{S \rightarrow T}$:
\begin{equation}
\mathcal{E}_{S \rightarrow T} = f_{\theta_e} (\mathbf{X}^{S}),
\end{equation}
which the decoder then transforms into target modality $\mathbf{X}^{T}$:
\begin{equation}
\mathbf{X}^{T} = f_{\theta_d} (\mathcal{E}_{S \rightarrow T}),
\end{equation}
following which the decoded modality $T$ is translated back into modality $S$:
\begin{align}
{\mathcal{E}}_{T \rightarrow S} = f_{\theta_e}(\hat{\mathbf{X}}^{T}_{}), \ \hat{\mathbf{X}}^{S}_{} = f_{\theta_d} ({\mathcal{E}}_{T \rightarrow S}).
\end{align}

The joint representation is learned by using a Seq2Seq model with attention~\citep{bahdanau2014neural} 
that translates source modality $\mathbf{X}^{S}$ to a target modality $\mathbf{X}^{T}$. While Seq2Seq models have been predominantly used for machine translation, we extend its usage to the realm of multimodal machine learning.

%
%
The hidden state output of each time step is based on the previous hidden state along with the input sequence and is constructed using a recurrent network.
\begin{equation}
\mathbf{h}_\ell = \mathtt{RNN}(\mathbf{h}_{\ell-1}, \mathbf{X}^{S}_\ell) \quad \forall \ell \in [1,L].
\end{equation}
The encoder's output is the concatenation of all hidden states of the encoding RNN,
\begin{equation}
\mathcal{E}_{S \rightarrow T} = [\mathbf{h}_{1}, \mathbf{h}_{2}, ..., \mathbf{h}_{L}],
\end{equation}
where $L$ is the length of the source modality $\mathbf{X}^{S}$.

The decoder maps the representation $\mathcal{E}_{S \rightarrow T}$ into the target modality $\mathbf{X}^{T}$. This is performed by decoding each token $\mathbf{X}^{T}_t$ at a time based on $\mathcal{E}_{S \rightarrow T}$ and all previous decoded tokens, which is formulated as
\begin{equation}
p(\mathbf{X}^{T}) = \prod_{\ell=1}^{L} p(\mathbf{X}^{T}_\ell|\mathcal{E}_{S \rightarrow T}, \mathbf{X}^{T}_1, ..., \mathbf{X}^{T}_{\ell-1}).
\end{equation}
\ours \ accepts variable-length inputs of $\mathbf{X}^{S}$ and $\mathbf{X}^{T}$, and is trained to maximize the translational condition probability $p(\mathbf{X}^{T}|\mathbf{X}^{S})$. The best translation sequence is then given by
\begin{equation}
\hat{\mathbf{X}^{T}} = \operatorname*{arg\,max}_{\mathbf{X}^{T}} p(\mathbf{X}^{T}|\mathbf{X}^{S}).
\end{equation}
%
We use the traditional beam search approach~\citep{seq2seq_nn} 
for decoding. 

To obtain the joint representation for multimodal prediction, we 
only use
the forward translated representation during inference 
to remove the 
dependency on the target modality at test time.
If cyclic translation is used, we denote the translated representation with the symbol $\leftrightarrows$: 
\begin{equation}
{\mathcal{E}}_{S \leftrightarrows T} = {\mathcal{E}}_{S \rightarrow T}.
\end{equation}
${\mathcal{E}}_{S \leftrightarrows T}$ is then used for sentiment prediction:
\begin{align}
\hat{\mathbf{y}}_{} &= g_{w}({\mathcal{E}}_{S \leftrightarrows T}).
\end{align}

\subsection{Coupled Translation-Prediction Objective}

Training is performed with paired multimodal data and labels in the training set $(\mathbf{X}^{S}_{tr},\mathbf{X}^{T}_{tr},\mathbf{y}_{tr})$ 
The first two losses are
the forward translation loss $\mathcal{L}_t$ defined as 
\begin{equation}
\mathcal{L}_t = \mathbb{E} [\ell_{\mathbf{X}^{T}}(\hat{\mathbf{X}}^{T}_{},{\mathbf{X}}^{T}_{})],
\label{lt}
\end{equation}
and the cycle consistency loss $\mathcal{L}_c$ defined as
\begin{equation}
\mathcal{L}_c = \mathbb{E} [\ell_{\mathbf{X}^{S}}(\hat{\mathbf{X}}^{S}_{},{\mathbf{X}}^{S}_{})]
\label{lc}
\end{equation}
where 
$\ell_{\mathbf{X}^{T}}$ and $\ell_{\mathbf{X}^{S}}$
represent the respective loss functions. 
We use the Mean Squared Error (MSE) between the ground-truth and translated modalities. Finally, the prediction loss $\mathcal{L}_p$ is defined as
\begin{align}
\mathcal{L}_p &= \mathbb{E} [ \ell_{\mathbf{y}}(\hat{\mathbf{y}}_{},\mathbf{y}_{})]
\label{lp}
\end{align}
with
a loss function $\ell_{\mathbf{y}}$ 
defined
over the labels.

Our \ours \ model is trained end-to-end with a coupled translation-prediction objective function defined as 
\begin{equation}
\mathcal{L} = \lambda_t \mathcal{L}_t + \lambda_c \mathcal{L}_c + \mathcal{L}_p.
\label{eq:objective}
\end{equation}
where $\lambda_t$, $\lambda_t$ are weighting hyperparameters. \ours \ parameters are learned by minimizing this objective function
\begin{align}
\theta_e^*, \theta_d^*, w^* &= \argmin_{\theta_e,\theta_d,w} \ [\lambda_t \mathcal{L}_t + \lambda_c \mathcal{L}_c + \mathcal{L}_p].
\end{align}

Parallel multimodal data is not required at test time. Inference is performed using only the source modality 
$\mathbf{X}^{S}$:
\begin{align}
{\mathcal{E}}_{S \leftrightarrows T} &= f_{\theta_e^*}
(\mathbf{X}^{S}), \\
\hat{\mathbf{y}} 
&= g_{w^*}({\mathcal{E}}_{S \leftrightarrows T}).
\end{align}
This is possible because the encoder $f_{\theta_e^*}$ has been trained to translate the source modality $\mathbf{X}^{S}$ into a joint representation ${\mathcal{E}}_{S \leftrightarrows T}$ that captures information from both source and target modalities. 

\subsection{Hierarchical \ours \ for Three Modalities}

\begin{figure}[tbp]
\centering
\includegraphics[width=0.5\textwidth]{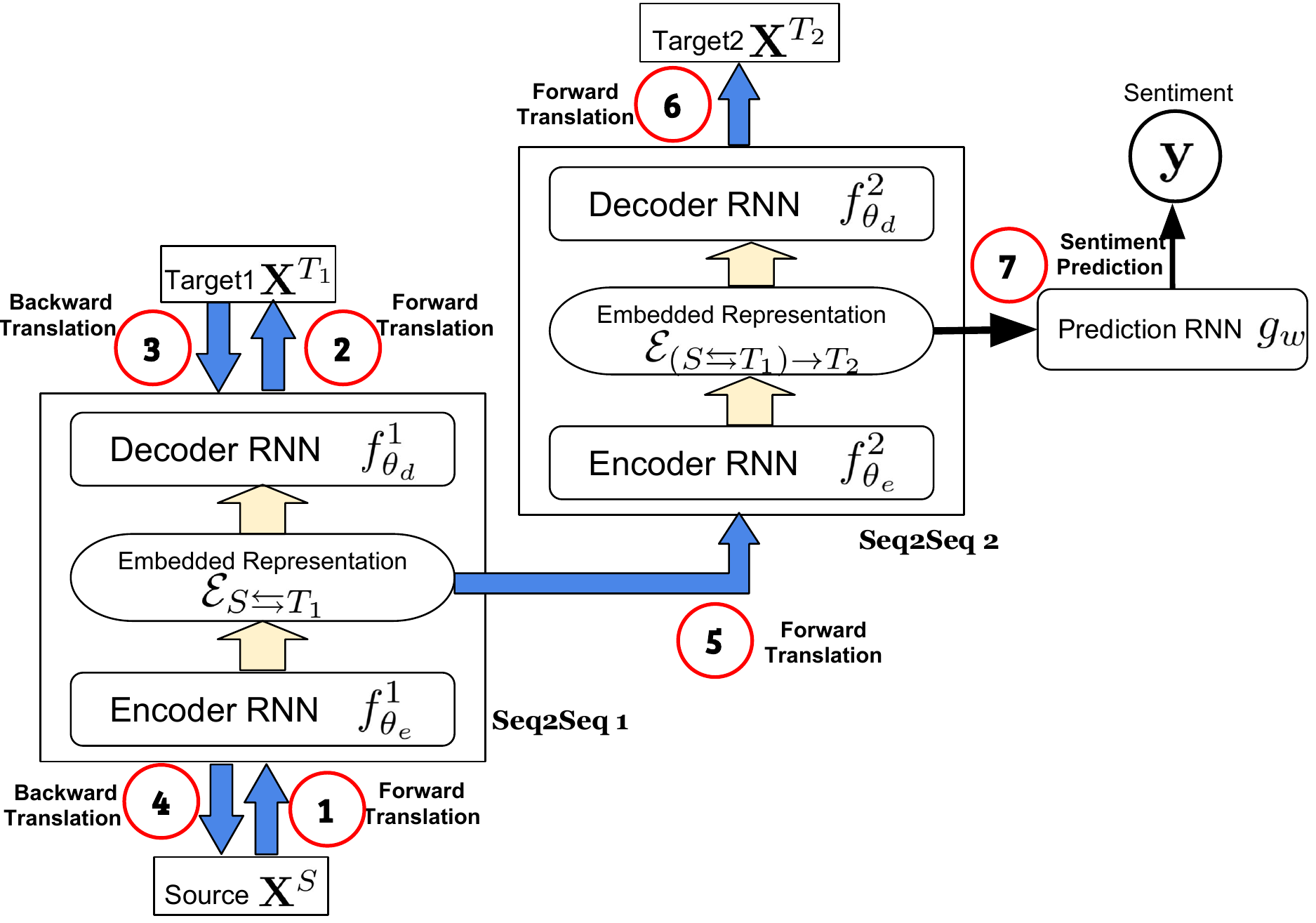}
\caption{
\small
{
Hierarchical \ours \ for three modalities: the source modality $\mathbf{X}^{S}$ and the target modalities $\mathbf{X}^{T_1}$ and $\mathbf{X}^{T_2}$. The joint representation ${\mathcal{E}}_{S \leftrightarrows T_1}$ is obtained via a cyclic translation between $\mathbf{X}^{S}$ and $\mathbf{X}^{T_1}$, then further translated into $\mathbf{X}^{T_2}$. Next, the joint representation of all three modalities, ${\mathcal{E}}_{(S \leftrightarrows T_1) \rightarrow T_2}$, is used for sentiment prediction. The model is trained end-to-end with a coupled translation-prediction objective. At test time, only the source modality $\mathbf{X}^{S}$ is required for prediction.}
}
\label{fig:h_s2s}
\end{figure}

We extend the \ours \ in a hierarchical manner to learn joint representations from more than two modalities. Figure~\ref{fig:h_s2s} shows the case for three modalities. The hierarchical \ours \ starts with a source modality $\mathbf{X}^{S}$ and two target modalities $\mathbf{X}^{T_1}$ and $\mathbf{X}^{T_2}$. To learn joint representations, two levels of modality translations are performed. The first level learns a joint representation from $\mathbf{X}^{S}$ and $\mathbf{X}^{T_1}$ using multimodal cyclic translations as defined previously. At the second level, a joint representation is learned hierarchically by translating the first representation $\mathcal{E}_{S \rightarrow T_1}$ into $\mathbf{X}^{T_2}$. For more than three modalities, the modality translation process can be repeated hierarchically.

Two Seq2Seq models are used in the hierarchical \ours \ for three modalities, denoted as encoder-decoder pairs $(f_{\theta_e}^1,f_{\theta_d}^1)$ and $(f_{\theta_e}^2,f_{\theta_d}^2)$. A multimodal cyclic translation is first performed between source modality $\mathbf{X}^{S}$ and the first target modality $\mathbf{X}^{T_1}$. The forward translation is defined as
\begin{align}
{\mathcal{E}}_{S \rightarrow T_1} = f_{\theta_e}^1 (\mathbf{X}^{S}_{tr}), \ \hat{\mathbf{X}}^{T_1}_{tr} = f_{\theta_d}^1 ({\mathcal{E}}_{S \rightarrow T_1}),
\end{align}
and followed by the decoded modality $\mathbf{X}^{T_1}$ being translated back into modality $\mathbf{X}^{S}$:
\begin{align}
{\mathcal{E}}_{T_1 \rightarrow S} = f_{\theta_e}^1(\hat{\mathbf{X}}^{T_1}_{tr}), \ \hat{\mathbf{X}}^{S}_{tr} = f_{\theta_d}^1 ({\mathcal{E}}_{T_1 \rightarrow S}).
\end{align}
A second hierarchical Seq2Seq model is applied on the outputs of the first encoder $f_{\theta_e}^1$:
\begin{align}
{\mathcal{E}}_{S \leftrightarrows T_1} &= {\mathcal{E}}_{S \rightarrow T_1}, \\
{\mathcal{E}}_{(S \leftrightarrows T_1) \rightarrow T_2} = f_{\theta_e}^2 ({\mathcal{E}}_{S \leftrightarrows T_1})&, \ \hat{\mathbf{X}}^{T_2}_{tr} = f_{\theta_d}^2 ({\mathcal{E}}_{(S \leftrightarrows T_1) \rightarrow T_2}).
\end{align}
The joint representation between modalities $\mathbf{X}^{S}$, $\mathbf{X}^{T_1}$ and $\mathbf{X}^{T_2}$ is now ${\mathcal{E}}_{(S \leftrightarrows T_1) \rightarrow T_2}$. 
It is used for sentiment prediction via a recurrent neural network 
via regression method.

Training the hierarchical \ours \ involves computing a cycle consistent loss for modality $T_1$, given by the respective forward translation loss $\mathcal{L}_{t_1}$ and the cycle consistency loss $\mathcal{L}_{c_1}$. We do not use a cyclic translation loss when translating from ${\mathcal{E}}_{S \leftrightarrows T_1}$ to $\mathbf{X}^{T_2}$ since the ground truth ${\mathcal{E}}_{S \leftrightarrows T_1}$ is unknown, and so only the translation loss $\mathcal{L}_{t_2}$ is computed. The final objective for hierarchical \ours \ is given by
\begin{align}
\mathcal{L} = \lambda_{t_1} \mathcal{L}_{t_1} + \lambda_{c_1} \mathcal{L}_{c_1} + \lambda_{t_2} \mathcal{L}_{t_2} + \mathcal{L}_p
\end{align}

We emphasize that for \ours \ with three modalities, \textit{only} a single source modality $\mathbf{X}^{S}$ is required at test time. Therefore, \ours \ has a significant advantage over existing models since it is robust to noisy or missing target modalities.

\section{Experimental Setup}\label{sec:Expset}

In this section, we describe our experimental methodology to evaluate the joint representations learned by \ours\footnote{Our source code is released at \url{https://github.com/hainow/MCTN}.}

\newcolumntype{K}[1]{>{\centering\arraybackslash}p{#1}}

\begin{table}[t!]
\fontsize{7.5}{10}\selectfont
\setlength\tabcolsep{1.0pt}
\begin{tabular}{l : c: *{4}{K{1.24cm}}}
\Xhline{3\arrayrulewidth} Dataset
& & \multicolumn{4}{c}{\textbf{CMU-MOSI}} \\
Model       & Test Inputs & Acc($\uparrow$) & F1($\uparrow$) & MAE($\downarrow$) & Corr($\uparrow$)\\ 
\Xhline{0.5\arrayrulewidth}
RF         & $\{\ell,v,a\}$ & 56.4 &  56.3  &	-   &  - \\ 
SVM	       & $\{\ell,v,a\}$ &71.6 &  72.3  & 1.100  &  0.559 \\ 
THMM	   & $\{\ell,v,a\}$ &50.7 &  45.4  & - & -\\
EF-HCRF		& $\{\ell,v,a\}$ &65.3 & 65.4 & - & -\\
MV-HCRF& $\{\ell,v,a\}$ &65.6 & 65.7 & - & -\\
DF              & $\{\ell,v,a\}$ &74.2 &   74.2   &  1.143     &  0.518 \\
EF-LSTM		& $\{\ell,v,a\}$ & 74.3 &   74.3   &  1.023  & 0.622 \\
MV-LSTM			& $\{\ell,v,a\}$ &73.9 &   74.0   & 1.019 & 0.601 \\
BC-LSTM         & $\{\ell,v,a\}$ &75.2 &   75.3   & 1.079 & 0.614 \\ 
TFN             & $\{\ell,v,a\}$ &74.6 &   74.5   & 1.040 & 0.587   \\ 
GME-LSTM(A) 	& $\{\ell,v,a\}$ &76.5 & 73.4 & 0.955 & - \\
MARN 			& $\{\ell,v,a\}$ &77.1 & 77.0 & 0.968 & 0.625 \\ 
MFN				& $\{\ell,v,a\}$ &77.4 & 77.3 & 0.965 & 0.632 \\
LMF     		& $\{\ell,v,a\}$ &76.4 & 75.7 & 0.912 & 0.668 \\
RMFN     		& $\{\ell,v,a\}$ &78.4 & 78.0 & 0.922 & \textbf{0.681} \\
\Xhline{0.5\arrayrulewidth}
{\ours}      	& $\{\ell\}$ &\textbf{79.3}	& \textbf{79.1}	& \textbf{0.909}  & 0.676 \\ 
\Xhline{3\arrayrulewidth}
\end{tabular}
\caption{
\small
{
Sentiment prediction results on CMU-MOSI. Best results are highlighted in bold. \ours \ outperforms the current state-of-the-art across most evaluation metrics and uses only the language modality during testing.}
}
\label{mosi}
\end{table}

\begin{table}[tb]
\fontsize{7.5}{10}\selectfont
\setlength\tabcolsep{1.0pt}
\begin{tabular}{l : c : *{2}{K{1.3cm}} : *{2}{K{1.3cm}}}
\Xhline{3\arrayrulewidth}
Dataset & & \multicolumn{2}{c:}{\textbf{ICT-MMMO}} & \multicolumn{2}{c}{\textbf{YouTube}} 
\\
Model        	& Test Inputs & Acc($\uparrow$) & F1($\uparrow$) & Acc($\uparrow$) & F1($\uparrow$) 
\\ 
\Xhline{0.5\arrayrulewidth}
RF				& $\{\ell,v,a\}$ & 70.0 & 69.8 & 33.3 & 32.3 
\\
SVM     		& $\{\ell,v,a\}$ &68.8 & 68.7 & 42.4 & 37.9 
\\
THMM			& $\{\ell,v,a\}$ &53.8	& 53.0 & 42.4 & 27.9 
\\
EF-HCRF	& $\{\ell,v,a\}$ &73.8 & 73.1 & 45.8 & 45.0 
\\
MV-HCRF	& $\{\ell,v,a\}$ &68.8 & 67.1 & 44.1 & 44.0
\\
DF   			& $\{\ell,v,a\}$ & 65.0	& 58.7 & 45.8 & 32.0 
\\
EF-LSTM	& $\{\ell,v,a\}$ &72.5 & 70.9 & 44.1 & 43.6 
\\
MV-LSTM			& $\{\ell,v,a\}$ &72.5 & 72.3 & 45.8 & 43.3 
\\
BC-LSTM    		& $\{\ell,v,a\}$ &70.0 & 70.1 & 45.0 & 45.1 
\\ 
TFN      		& $\{\ell,v,a\}$ &72.5 & 72.6 & 45.0 & 41.0 
\\ 
MARN			& $\{\ell,v,a\}$ &71.3 & 70.2 & 48.3 & 44.9 
\\ 
MFN				& $\{\ell,v,a\}$ &73.8 & 73.1 & \textbf{51.7} & 51.6 
\\
\Xhline{0.5\arrayrulewidth}
{\ours} & $\{\ell\}$
& \textbf{81.3} & \textbf{80.8}
& \textbf{51.7}	& \textbf{52.4} 
\\ 
\Xhline{0.5\arrayrulewidth}
\Xhline{3\arrayrulewidth}
\end{tabular}
\caption{
\small
{Sentiment prediction results on ICT-MMMO and YouTube. Best results are highlighted in bold. \ours \ outperforms the current state-of-the-art across most evaluation metrics and uses only the language modality during testing.}
}
\label{full}
\end{table}

\subsection{Dataset and Input Modalities}
\label{sec:MOSI}
We use the CMU Multimodal Opinion-level Sentiment Intensity dataset (CMU-MOSI) which contains 2199 video segments each with a sentiment label in the range $[-3,+3]$. To be consistent with prior work, we use 52 segments for training, 10 for validation and 31 for testing. The same speaker does not appear in both training and testing sets to ensure that our model learns speaker-independent representations. We also run experiments on ICT-MMMO~\citep{wollmer2013youtube} and YouTube~\citep{morency2011towards} which consist of online review videos annotated for sentiment.

\subsection{Multimodal Features and Alignment}
Following previous work~\citep{multistage}, GloVe word embeddings~\citep{pennington2014glove}, Facet~\citep{emotient}, and COVAREP~\citep{degottex2014covarep} features are extracted for the language, visual and acoustic modalities respectively\footnote{Details on feature extraction are in supplementary material.}. Forced alignment is performed using P2FA~\citep{P2FA} to obtain spoken word utterance times. The visual and acoustic features are aligned by computing their average over the utterance interval of each word.

\subsection{Evaluation Metrics}
For parameter optimization on CMU-MOSI, the prediction loss function is set as the Mean Absolute Error (MAE): $\ell_p (\hat{\mathbf{y}}_{train},\mathbf{y}_{train}) = |\hat{\mathbf{y}}_{train}-\mathbf{y}_{train}|$. We report MAE and Pearson's correlation $r$. We also perform sentiment classification on CMU-MOSI and report binary accuracy (Acc) and F1 score (F1). On ICT-MMMO and YouTube, we set the prediction loss function as categorical cross-entropy and report sentiment classification and F1 score. For all metrics, higher values indicate stronger performance, except MAE where lower values indicate stronger performance. 

\subsection{Baseline Models}
\label{sec:base}

\begin{figure*}[ht]
\centering
\includegraphics[width=\textwidth]{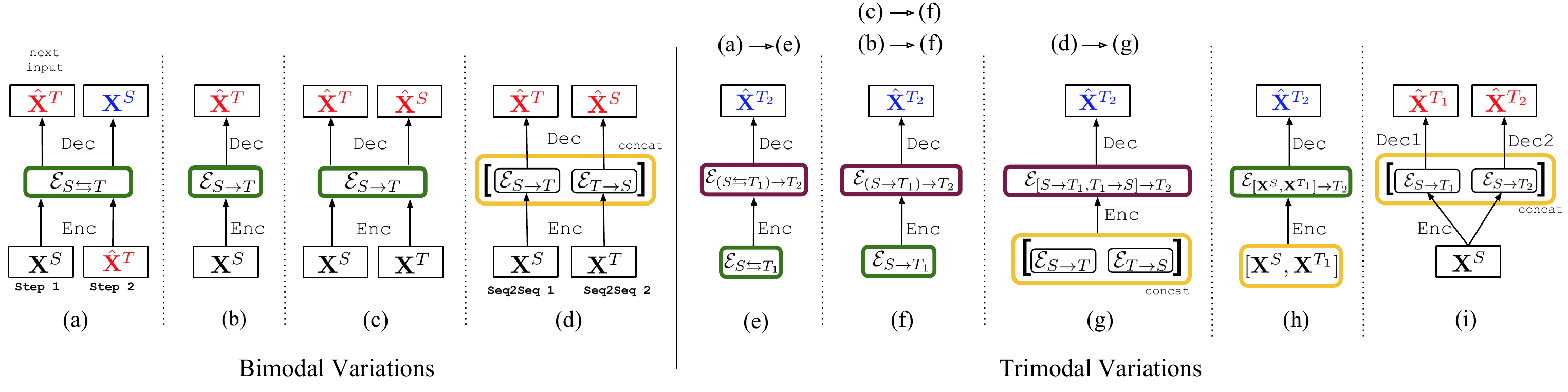}
\caption{
\small
{Variations of our models: 
(a) \ours \ Bimodal with cyclic translation, 
(b) Simple Bimodal without cyclic translation, 
(c) No-Cycle Bimodal with different inputs of the same modality pair, and without cyclic translation, 
(d) Double Bimodal for two modalities without cyclic translation, with two different inputs (of the same pair), 
(e) \ours \ Trimodal with input from (a),
(f) Simple Trimodal for three modalities, with input as a joint representation taken from previous \ours \ for two modalities from (b) or (c), 
(g) Double Trimodal with input from (d), 
(h) Concat Trimodal which is similar to (b) but with input as the concatenation of 2 modalities, 
(i) Paired Trimodal using one encoder and 2 separate decoders for modality translations.
\textit{Legend}: black modality is ground truth, red (``hat'') modality represents translated output, blue (``hat'') modality is target output from previous translation outputs, and yellow box denotes concatenation.}
}
\label{fig:variants}
\end{figure*}
\addtocounter{figure}{-1}

We compare to the following multimodal models: \textit{RMFN}~\citep{multistage} uses a multistage approach to learn hierarchical representations (current state-of-the-art on CMU-MOSI). \textit{LMF}~\citep{lowrank} approximates the expensive tensor products in \textit{TFN}~\citep{tensoremnlp17} with efficient low-rank factors. \textit{MFN}~\citep{zadeh2018memory} synchronizes sequences using a multimodal gated memory. \textit{EF-LSTM} concatenates multimodal inputs and uses a single LSTM~\citep{Hochreiter:1997:LSM:1246443.1246450}. For a description of other baselines, please refer to the supplementary material.

\section{Results and Discussion} \label{sec:Results}

This section presents and discusses our experimental results.

\subsection{Comparison with Existing Work}

\textit{Q1: 
How does \ours \ compare with existing state-of-the-art approaching for multimodal sentiment analysis?}

We compare \ours \ with  previous models \footnote{For full results please refer to the supplementary material.}. From Table~\ref{mosi}, \ours \ using language as the source modality achieves new start-of-the-art results on CMU-MOSI for multimodal sentiment analysis. State-of-the-art results are also achieved on ICT-MMMO and YouTube (Table~\ref{full}). It is important to note that \ours \ only uses language during testing, while other baselines use all three modalities.

\subsection{Adding More Modalities}

\textit{Q2: What is the impact of increasing the number of modalities during training for \ours \ with cyclic translations?}

\begin{figure}[!tb]
\centering
\setlength\tabcolsep{1.0pt}
\begin{tabular}{ccc}
\begin{subfigure}{0.33\linewidth}\centering\includegraphics[width=\columnwidth]{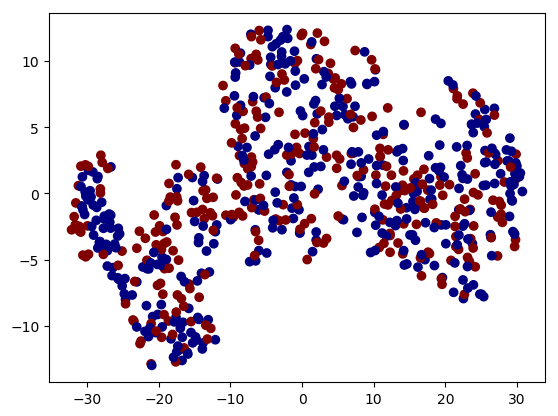}\caption{
\centering
\footnotesize{
\centering
\ours \ Bimodal \textit{without} cyclic translations}
}\label{fig:taba}\end{subfigure} &
\begin{subfigure}{0.33\linewidth}\centering\includegraphics[width=\columnwidth]{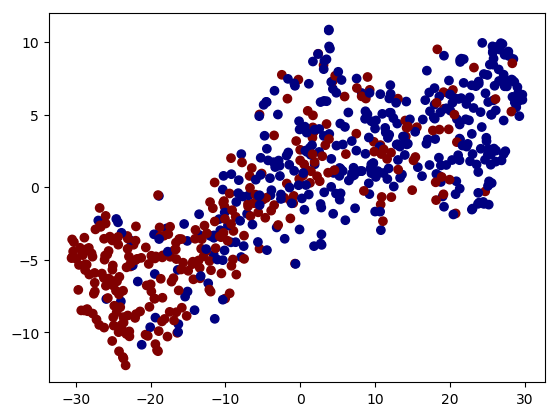}\caption{
\centering
\footnotesize{\ours \ Bimodal \textit{with} cyclic translations}
}\label{fig:tabb}\end{subfigure} &
\begin{subfigure}{0.33\linewidth}\centering\includegraphics[width=\columnwidth]{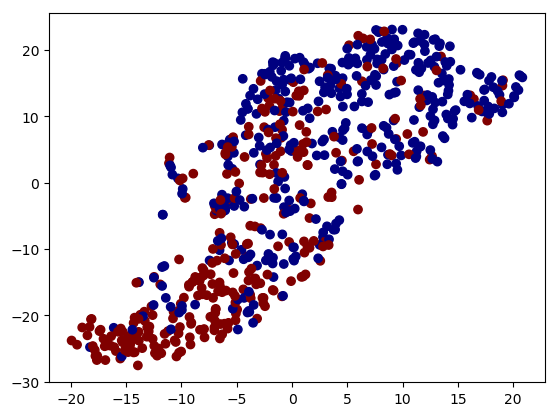}\caption{
\centering
\footnotesize{\ours \ Trimodal \textit{with} cyclic translations}
}\label{fig:tabc}\end{subfigure}\\
\end{tabular}
\captionof{figure}[]{
\small
{t-SNE visualization of the joint representations learned by \ours. \textit{Legend}: red: videos with negative sentiment, blue: videos with positive sentiment. Adding modalities and using cyclic translations improve discriminative performance and leads to increasingly separable representations.}
}
\label{fig:tsne}
\end{figure}

\begin{table}[t!]
\fontsize{7.5}{10}\selectfont
\setlength\tabcolsep{1.0pt}
\begin{tabular}{l : *{16}{K{0.95cm}}}
\Xhline{3\arrayrulewidth}
Dataset & \multicolumn{5}{c}{\textbf{CMU-MOSI}} \\
Model       & \multicolumn{1}{c}{Translation} & Acc & F1 & MAE & Corr\\ 
\Xhline{0.5\arrayrulewidth}
\multirow{3}{*}{\ours \ Bimodal (\ref{fig:variants}a)} 
& \multicolumn{1}{c}{$V \leftrightarrows A$} & \multicolumn{1}{c}{{53.1}} & \multicolumn{1}{c}{{53.2}} & \multicolumn{1}{c}{{1.420}} & \multicolumn{1}{c}{{0.034}}  \\
& \multicolumn{1}{c}{$T \leftrightarrows A$} & \multicolumn{1}{c}{{76.4}} & \multicolumn{1}{c}{{76.4}} & \multicolumn{1}{c}{{0.977}} & \multicolumn{1}{c}{{0.636}}  \\
& \multicolumn{1}{c}{$T \leftrightarrows V$} & \multicolumn{1}{c}{{76.8}} & \multicolumn{1}{c}{{76.8}} & \multicolumn{1}{c}{{1.034}} & \multicolumn{1}{c}{{0.592}}  \\

\Xhline{0.5\arrayrulewidth}

\multirow{3}{*}{\ours \ Trimodal (\ref{fig:variants}e)} 
& \multicolumn{1}{c}{$(V \leftrightarrows A) \rightarrow T$} & \multicolumn{1}{c}{{56.4}} & \multicolumn{1}{c}{{56.3}} & \multicolumn{1}{c}{{1.455}} & \multicolumn{1}{c}{{0.151}}  \\
& \multicolumn{1}{c}{$(T \leftrightarrows A) \rightarrow V$} & \multicolumn{1}{c}{{78.7}} & \multicolumn{1}{c}{{78.8}} & \multicolumn{1}{c}{{0.960}} & \multicolumn{1}{c}{{0.650}}  \\
& \multicolumn{1}{c}{$(T \leftrightarrows V) \rightarrow A$} & \multicolumn{1}{c}{\textbf{79.3}} & \multicolumn{1}{c}{\textbf{79.1}} & \multicolumn{1}{c}{\textbf{0.909}} & \multicolumn{1}{c}{\textbf{0.676}}  \\
\Xhline{3\arrayrulewidth}
\end{tabular}
\caption{
\small
{
\ours \ performance improves as more modalities are introduced for cyclic translations during training.}
}
\label{tbl:trimodal}
\end{table}

We run experiments with \ours \ using combinations of two or three modalities with cyclic translations. From Table~\ref{tbl:trimodal}, we observe that adding more modalities improves performance, indicating that the joint representations learned are leveraging the information from more input modalities. This also implies that cyclic translations are a viable method to learn joint representations from multiple modalities since little information is lost from adding more modality translations. Another observation is that using language as the source modality always leads to the best performance, which is intuitive since the language modality contains the most discriminative information for sentiment~\citep{tensoremnlp17}. 

In addition, we visually inspect the joint representations learned from \ours \ as we add more modalities during training
(see Table~\ref{tbl:ablation}). 
The joint representations for each segment in CMU-MOSI are extracted from the best performing model for each number of modalities and then projected into two dimensions via the t-SNE algorithm~\citep{vanDerMaaten2008}. Each point is colored red or blue depending on whether the video segment is annotated for positive or negative sentiment. From Figure~\ref{fig:tsne}, we observe that the joint representations become increasingly separable as the more modalities are added when the \ours \ is trained. This is consistent with increasing discriminative performance with more modalities (as seen in Table~\ref{tbl:trimodal}).




\subsection{Ablation Studies}

\begin{table}[t!]
\fontsize{7.5}{10}\selectfont
\setlength\tabcolsep{1.0pt}
\begin{tabular}{l : *{16}{K{0.85cm}}}
\Xhline{3\arrayrulewidth}
Dataset & \multicolumn{5}{c}{\textbf{CMU-MOSI}} \\
Model       & \multicolumn{1}{c}{Translation} & Acc($\uparrow$) & F1($\uparrow$) & MAE($\downarrow$) & Corr($\uparrow$)\\ 
\Xhline{0.5\arrayrulewidth}
\multirow{3}{*}{\ours \ Bimodal (\ref{fig:variants}a)} 
& \multicolumn{1}{c}{$V \leftrightarrows A$} & \multicolumn{1}{c}{{53.1}} & \multicolumn{1}{c}{{53.2}} & \multicolumn{1}{c}{{1.420}} & \multicolumn{1}{c}{{0.034}}  \\
& \multicolumn{1}{c}{$T \leftrightarrows A$} & \multicolumn{1}{c}{{76.4}} & \multicolumn{1}{c}{{76.4}} & \multicolumn{1}{c}{{0.977}} & \multicolumn{1}{c}{\textbf{0.636}}  \\
& \multicolumn{1}{c}{$T \leftrightarrows V$} & \multicolumn{1}{c}{\textbf{76.8}} & \multicolumn{1}{c}{\textbf{76.8}} & \multicolumn{1}{c}{{1.034}} & \multicolumn{1}{c}{{0.592}}  \\
\Xhline{0.5\arrayrulewidth}
\multirow{3}{*}{Simple Bimodal (\ref{fig:variants}b)} 
& \multicolumn{1}{c}{$V \rightarrow A$} & \multicolumn{1}{c}{{55.4}} & \multicolumn{1}{c}{{55.5}} & \multicolumn{1}{c}{{1.422}} & \multicolumn{1}{c}{{0.119}}  \\
& \multicolumn{1}{c}{$T \rightarrow A$} & \multicolumn{1}{c}{{74.2}} & \multicolumn{1}{c}{{74.2}} & \multicolumn{1}{c}{{0.988}} & \multicolumn{1}{c}{{0.616}}  \\
& \multicolumn{1}{c}{$T \rightarrow V$} & \multicolumn{1}{c}{{75.7}} & \multicolumn{1}{c}{{75.6}} & \multicolumn{1}{c}{{1.002}} & \multicolumn{1}{c}{{0.617}}  \\
\Xhline{0.5\arrayrulewidth}
\multirow{3}{*}{No-Cycle Bimodal (\ref{fig:variants}c)} 
& \multicolumn{1}{c}{$V \rightarrow A,  \, A \rightarrow V$} & \multicolumn{1}{c}{{55.4}} & \multicolumn{1}{c}{{55.5}} & \multicolumn{1}{c}{{1.422}} & \multicolumn{1}{c}{{0.119}}  \\
& \multicolumn{1}{c}{$T \rightarrow A,  \, A \rightarrow T$} & \multicolumn{1}{c}{{75.5}} & \multicolumn{1}{c}{{75.6}} & \multicolumn{1}{c}{\textbf{0.971}} & \multicolumn{1}{c}{{0.629}}  \\
& \multicolumn{1}{c}{$T \rightarrow V,  \, V \rightarrow T$} & \multicolumn{1}{c}{{75.2}} & \multicolumn{1}{c}{{75.3}} & \multicolumn{1}{c}{{0.972}} & \multicolumn{1}{c}{{0.627}}  \\
\Xhline{0.5\arrayrulewidth}
\multirow{4}{*}{Double Bimodal (\ref{fig:variants}d)} 
& \multicolumn{1}{c}{$[V \rightarrow A, A \rightarrow V]$} & \multicolumn{1}{c}{{57.0}} & \multicolumn{1}{c}{{57.1}} & \multicolumn{1}{c}{{1.502}} & \multicolumn{1}{c}{{0.168}}  \\
& \multicolumn{1}{c}{$[T \rightarrow A, A \rightarrow T]$} & \multicolumn{1}{c}{{72.3}} & \multicolumn{1}{c}{{72.3}} & \multicolumn{1}{c}{{1.035}} & \multicolumn{1}{c}{{0.578}}  \\
& \multicolumn{1}{c}{$[T \rightarrow V, V \rightarrow T]$} & \multicolumn{1}{c}{{73.3}} & \multicolumn{1}{c}{{73.4}} & \multicolumn{1}{c}{{1.020}} & \multicolumn{1}{c}{{0.570}}  \\
\Xhline{3\arrayrulewidth}
\end{tabular}
\caption{
\small
{
Bimodal variations results on CMU-MOSI dataset. \ours \ Bimodal with cyclic translations performs best.}
}

\label{tbl:baselines}
\end{table}

\begin{table}[t!]
\fontsize{7.5}{10}\selectfont
\setlength\tabcolsep{1.0pt}
\begin{tabular}{l : *{16}{K{0.75cm}}}
\Xhline{3\arrayrulewidth}
Dataset & \multicolumn{5}{c}{\textbf{CMU-MOSI}} \\
Model       & \multicolumn{1}{c}{Translation} & Acc($\uparrow$) & F1($\uparrow$) & MAE($\downarrow$) & Corr($\uparrow$)\\ 
\Xhline{0.5\arrayrulewidth}
\multirow{3}{*}{\ours \ Trimodal (\ref{fig:variants}e)} 
& \multicolumn{1}{c}{$(V \leftrightarrows A) \rightarrow T$} & \multicolumn{1}{c}{{56.4}} & \multicolumn{1}{c}{{56.3}} & \multicolumn{1}{c}{{1.455}} & \multicolumn{1}{c}{{0.151}}  \\
%
& \multicolumn{1}{c}{$(T \leftrightarrows A) \rightarrow V$} & \multicolumn{1}{c}{{78.7}} & \multicolumn{1}{c}{{78.8}} & \multicolumn{1}{c}{{0.960}} & \multicolumn{1}{c}{{0.650}}  \\
& \multicolumn{1}{c}{$(T \leftrightarrows V) \rightarrow A$} & \multicolumn{1}{c}{\textbf{79.3}} & \multicolumn{1}{c}{\textbf{79.1}} & \multicolumn{1}{c}{\textbf{0.909}} & \multicolumn{1}{c}{\textbf{0.676}}  \\
\Xhline{0.5\arrayrulewidth}
\multirow{6}{*}{Simple Trimodal (\ref{fig:variants}f)} 
& \multicolumn{1}{c}{$(V \rightarrow T) \rightarrow A$} & \multicolumn{1}{c}{{54.1}} & \multicolumn{1}{c}{{52.9}} & \multicolumn{1}{c}{{1.408}} & \multicolumn{1}{c}{{0.040}}  \\
& \multicolumn{1}{c}{$(V \rightarrow A) \rightarrow T$} & \multicolumn{1}{c}{{52.0}} & \multicolumn{1}{c}{{51.9}} & \multicolumn{1}{c}{{1.439}} & \multicolumn{1}{c}{{0.015}}  \\
& \multicolumn{1}{c}{$(A \rightarrow V) \rightarrow T$} & \multicolumn{1}{c}{{56.6}} & \multicolumn{1}{c}{{56.7}} & \multicolumn{1}{c}{{1.593}} & \multicolumn{1}{c}{{0.067}}  \\
& \multicolumn{1}{c}{$(A \rightarrow T) \rightarrow V$} & \multicolumn{1}{c}{{54.1}} & \multicolumn{1}{c}{{54.2}} & \multicolumn{1}{c}{{1.577}} & \multicolumn{1}{c}{{0.028}}  \\
& \multicolumn{1}{c}{$(T \rightarrow A) \rightarrow V$} & \multicolumn{1}{c}{{74.3}} & \multicolumn{1}{c}{{74.4}} & \multicolumn{1}{c}{{1.001}} & \multicolumn{1}{c}{{0.609}}  \\
& \multicolumn{1}{c}{$(T \rightarrow V) \rightarrow A$} & \multicolumn{1}{c}{{74.3}} & \multicolumn{1}{c}{{74.4}} & \multicolumn{1}{c}{{0.997}} & \multicolumn{1}{c}{{0.596}}  \\
\Xhline{0.5\arrayrulewidth}
\multirow{1}{*}{Double Trimodal (\ref{fig:variants}g)} 
& \multicolumn{1}{c}{$[T \rightarrow V,  \, V \rightarrow T] \rightarrow A$} & \multicolumn{1}{c}{{73.3}} & \multicolumn{1}{c}{{73.1}} & \multicolumn{1}{c}{{1.058}} & \multicolumn{1}{c}{{0.578}}  \\
\Xhline{0.5\arrayrulewidth}
\multirow{7}{*}{Concat Trimodal (\ref{fig:variants}h)} 
& \multicolumn{1}{c}{$[V, A] \rightarrow T$} & \multicolumn{1}{c}{{55.0}} & \multicolumn{1}{c}{{54.6}} & \multicolumn{1}{c}{{1.535}} & \multicolumn{1}{c}{{0.176}}  \\
& \multicolumn{1}{c}{$[A, T] \rightarrow V$} & \multicolumn{1}{c}{{73.3}} & \multicolumn{1}{c}{{73.4}} & \multicolumn{1}{c}{{1.060}} & \multicolumn{1}{c}{{0.561}}  \\
& \multicolumn{1}{c}{$[T, V] \rightarrow A$} & \multicolumn{1}{c}{{72.3}} & \multicolumn{1}{c}{{72.3}} & \multicolumn{1}{c}{{1.068}} & \multicolumn{1}{c}{{0.576}}  \\
& \multicolumn{1}{c}{$A \rightarrow [T, V]$} & \multicolumn{1}{c}{{55.5}} & \multicolumn{1}{c}{{55.6}} & \multicolumn{1}{c}{{1.617}} & \multicolumn{1}{c}{{0.056}}  \\
& \multicolumn{1}{c}{$T \rightarrow [A, V]$} & \multicolumn{1}{c}{{75.7}} & \multicolumn{1}{c}{{75.7}} & \multicolumn{1}{c}{{0.958}} & \multicolumn{1}{c}{{0.634}}  \\
& \multicolumn{1}{c}{$[T, A] \rightarrow [T, V]$} & \multicolumn{1}{c}{{73.2}} & \multicolumn{1}{c}{{73.2}} & \multicolumn{1}{c}{{1.008}} & \multicolumn{1}{c}{{0.591}}  \\
& \multicolumn{1}{c}{$[T, V] \rightarrow [T, A]$} & \multicolumn{1}{c}{{74.1}} & \multicolumn{1}{c}{{74.1}} & \multicolumn{1}{c}{{0.999}} & \multicolumn{1}{c}{{0.607}}  \\
\Xhline{0.5\arrayrulewidth}
Paired Trimodal (\ref{fig:variants}i) & 
\multicolumn{1}{c}{$[T \rightarrow A, T \rightarrow V]$} & 73.8 & 73.8 & 1.022 & 0.611 \\
\Xhline{3\arrayrulewidth}
\end{tabular}
\caption{
\small
{
Trimodal variations results on CMU-MOSI dataset. \ours \ (hierarchical) with cyclic translations performs best.}
}
\label{tbl:ablation}
\end{table}

We use several models to test our design decisions. Specifically, we evaluate the impact of cyclic translations, modality ordering, and hierarchical structure.

For bimodal \ours, we design the following ablation models shown in the left half of Figure~\ref{fig:variants}: (a) \ours \ bimodal between $\mathbf{X}^S$ and $\mathbf{X}^T$, (b) simple bimodal by translating from $\mathbf{X}^S$ to $\mathbf{X}^T$ without cyclic loss, (c) no-cycle bimodal which does not use cyclic translations but rather performs two independent translations between $\mathbf{X}^S$ and $\mathbf{X}^T$, (d) double bimodal: two seq2seq models with different inputs (of the same modality pair) and then  using the concatenation of the joint representations ${\mathcal{E}}_{S \rightarrow T}$ and ${\mathcal{E}}_{T \rightarrow S}$ as the final embeddings. 
%

For trimodal \ours, we design the following ablation models shown in the right half of Figure~\ref{fig:variants}: (e) \ours \ trimodal which uses the proposed hierarchical translations between $\mathbf{X}^S$, $\mathbf{X}^{T_1}$ and $\mathbf{X}^{T_2}$, (f) simple trimodal based on translation from $\mathbf{X}^S$ to $\mathbf{X}^{T_1}$ without cyclic translations, (g) double trimodal extended from (d) which does not use cyclic translations but rather performs two independent translations between $\mathbf{X}^S$ and $\mathbf{X}^{T_1}$, (h) concat trimodal which does not perform a first level of cyclic translation but directly translates the concatenated modality pair $[\mathbf{X}^S, \mathbf{X}^{T_1}]$ into $\mathbf{X}^{T_2}$, and finally, (i) paired trimodal which uses two separate decoders on top of the intermediate representation.

\textit{Q3: What is the impact of cyclic translations in \ours?}

The bimodal results are in Table~\ref{tbl:baselines}. The models that employ cyclic translations (Figure~\ref{fig:variants}(a)) outperform all other models. The trimodal results are in Table~\ref{tbl:ablation} and we make a similar observation: Figure~\ref{fig:variants}(e) with cyclic translations outperforms the baselines (f), (g) and (h). The gap for the trimodal case is especially large. This implies that using cyclic translations is crucial for learning discriminative joint representations. Our intuition is that using cyclic translations: (1) encourages the model to enforce symmetry between the representations from source and target modalities thus adding a source of regularization, and (2) ensures that the representation retains maximal information from all modalities.






\textit{Q4: What is the effect of using two Seq2Seq models instead of one shared Seq2Seq model for cyclic translations?}

We compare Figure~\ref{fig:variants}(c), which uses one Seq2Seq model for cyclic translations with Figure~\ref{fig:variants}(d), which uses two separate Seq2Seq models: one for forward translation and one for backward translation. We observe from Table~\ref{tbl:baselines} that (c) $>$ (d), so using one model with shared parameters is better. This is also true for hierarchical \ours: (f) $>$ (g) in Table~\ref{tbl:ablation}. We hypothesize that this is because training two deep Seq2Seq models requires more data and is prone to overfitting. 
Also, it does not learn only a single joint representation but instead two separate representations. 

\textit{Q5: What is the impact of varying source and target modalities for cyclic translations?}

From Tables \ref{tbl:trimodal}, \ref{tbl:baselines} and \ref{tbl:ablation}, we observe that language contributes most towards the joint representations. For bimodal cases, combining language with visual is generally better than combining the language and acoustic modalities. For hierarchical \ours, presenting language as the source modality leads to the best performance, and a first level of cyclic translations between language and visual is better than between language and audio. On the other hand, only translating between visual and acoustic modalities dramatically decreases performance. Further adding language as a target modality for hierarchical \ours \ will not help much as well. Overall, for the \ours, language appears to be the most discriminative modality making it crucial to be used as the source modality during translations.

\textit{Q6: What is the impact of using two levels of translations instead of one level when learning from three modalities?}

Our hierarchical \ours \ is shown in Figure~\ref{fig:variants}(e). In Figure~\ref{fig:variants}(h), we concatenate two modalities as input and use only one phase of translation. From Table \ref{tbl:ablation}, we observe that (e) $>$ (h): both levels of modality translations are important in the hierarchical \ours. We believe that representation learning is easier when the task is broken down recursively: using two translations each between a single pair of modalities, rather than a single translation between all modalities.

\section{Conclusion}
This paper investigated learning joint representations via cyclic translations from source to target modalities. During testing, we only need the source modality for prediction which ensures robustness to noisy or missing target modalities. We demonstrate that cyclic translations and seq2seq models are useful for learning joint representations in multimodal environments. In addition to achieving new state-of-the-art results on three datasets, our model learns increasingly discriminative joint representations with more input modalities while maintaining robustness to all target modalities.

\section{Acknowledgements}
PPL and LPM are partially supported by the NSF (Award \#1833355) and Oculus VR. HP and BP are supported by NSF grant IIS1563887 and the DARPA
D3M program. Any opinions, findings, and conclusions or recommendations expressed in this material are those of the author(s) and do not necessarily reflect the views of National Science Foundation, DARPA, or Oculus VR, and no official endorsement should be inferred. The authors thank Hieu Pham, Amir Zadeh, and anonymous reviewers for useful discussions and feedback.

\small
\bibliography{main}
\bibliographystyle{aaai}

\clearpage

\appendix

\section{Multimodal Features}

Here we present extra details on feature extraction for the language, visual and acoustic modalities.

\noindent \textbf{Language:} We used 300 dimensional Glove word embeddings trained on 840 billion tokens from the common crawl dataset~\citep{pennington2014glove}. These word embeddings were used to embed a sequence of individual words from video segment transcripts into a sequence of word vectors that represent spoken text. 

\noindent \textbf{Visual:} The library Facet~\citep{emotient} is used to extract a set of visual features including facial action units, facial landmarks, head pose, gaze tracking and HOG features~\citep{zhu2006fast}. These visual features are extracted from the full video segment at 30Hz to form a sequence of facial gesture measures throughout time.

\noindent \textbf{Acoustic:} The software COVAREP~\citep{degottex2014covarep} is used to extract acoustic features including 12 Mel-frequency cepstral coefficients, pitch tracking and voiced/unvoiced segmenting features~\citep{drugman2011joint}, glottal source parameters~\citep{childers1991vocal,drugman2012detection,alku1992glottal,alku1997parabolic,alku2002normalized}, peak slope parameters and maxima dispersion quotients~\citep{kane2013wavelet}. These visual features are extracted from the full audio clip of each segment at 100Hz to form a sequence that represent variations in tone of voice over an audio segment.

\section{Multimodal Alignment}
We perform forced alignment using P2FA~\citep{P2FA} to obtain the exact utterance time-stamp of each word. This allows us to align the three modalities together. Since words are considered the basic units of language we use the interval duration of each word utterance as one time-step. We acquire the aligned video and audio features by computing the expectation of their modality feature values over the word utterance time interval~\citep{multistage}.

\subsection{Baseline Models}

We also implement the Stacked, (\textit{EF-SLSTM})~\citep{6638947}, Bidirectional (\textit{EF-BLSTM})~\citep{Schuster:1997:BRN:2198065.2205129}, and Stacked Bidirectional (\textit{EF-SBLSTM}) LSTMs, as well as the following baselines: \textit{BC-LSTM}~\citep{contextmultimodalacl2017}, \textit{EF-HCRF}~\citep{Quattoni:2007:HCR:1313053.1313265}, \textit{EF/MV-LDHCRF}~\citep{morency2007latent}, \textit{MV-HCRF}~\citep{song2012multi}, \textit{EF/MV-HSSHCRF}~\citep{song2013action}, \textit{MV-LSTM}~\citep{rajagopalan2016extending}, \textit{DF}~\citep{Nojavanasghari:2016:DMF:2993148.2993176}, \textit{SAL-CNN}~\citep{wang2016select}, \textit{C-MKL}~\citep{poria2015deep}, \textit{THMM}~\citep{morency2011towards}, \textit{SVM}~\citep{cortes1995support,Park:2014:CAP:2663204.2663260} and \textit{RF}~\citep{Breiman:2001:RF:570181.570182}.

\section{Full Results}

We present the full results across all baseline models in Table~\ref{mosi_supp} and Table~\ref{full_supp}. \ours \ using all modalities achieves new start-of-the-art results on binary classification accuracy, F1 score, and MAE on the CMU-MOSI dataset for multimodal sentiment analysis. State-of-the-art results are also achieved on the ICT-MMMO and YouTube datasets (Table~\ref{full_supp}). These results are even more impressive considering that \ours \ only uses the language modality during testing, while other baseline models use all three modalities.

\newcolumntype{K}[1]{>{\centering\arraybackslash}p{#1}}

\begin{table}[t!]
\fontsize{7.5}{10}\selectfont
\setlength\tabcolsep{1.0pt}
\begin{tabular}{l : c: *{4}{K{1.24cm}}}
\Xhline{3\arrayrulewidth} Dataset
& & \multicolumn{4}{c}{\textbf{CMU-MOSI}} \\
Model       & Test Inputs & Acc($\uparrow$) & F1($\uparrow$) & MAE($\downarrow$) & Corr($\uparrow$)\\ 
\Xhline{0.5\arrayrulewidth}
RF         & $\{\ell,v,a\}$ & 56.4 &  56.3  &	-   &  - \\ 
SVM	       & $\{\ell,v,a\}$ &71.6 &  72.3  & 1.100  &  0.559 \\ 
THMM	   & $\{\ell,v,a\}$ &50.7 &  45.4  & - & -\\
EF-HCRF		& $\{\ell,v,a\}$& 65.3 & 65.4 & - & -\\
EF-LDHCRF		& $\{\ell,v,a\}$& 64.0 & 64.0 & - & -\\
MV-HCRF		& $\{\ell,v,a\}$& 44.8 & 27.7 & - & -\\
MV-LDHCRF		& $\{\ell,v,a\}$& 64.0 & 64.0 & - & -\\
CMV-HCRF		& $\{\ell,v,a\}$& 44.8 & 27.7 & - & -\\
CMV-LDHCRF		& $\{\ell,v,a\}$& 63.6 & 63.6 & - & -\\
EF-HSSHCRF		& $\{\ell,v,a\}$& 63.3 & 63.4 & - & -\\
MV-HSSHCRF		& $\{\ell,v,a\}$& 65.6 & 65.7 & - & -\\
DF              & $\{\ell,v,a\}$ &74.2 &   74.2   &  1.143     &  0.518 \\
EF-LSTM        & $\{\ell,v,a\}$& 74.3 &   74.3   &  1.023     &  0.622 \\
EF-SLSTM		& $\{\ell,v,a\}$& 72.7 & 72.8 & 1.081 & 0.600 \\
EF-BLSTM		& $\{\ell,v,a\}$& 72.0 & 72.0 &  1.080 & 0.577 \\
EF-SBLSTM		& $\{\ell,v,a\}$& 73.3 &   73.2    & 1.037 & 0.619 \\
MV-LSTM			& $\{\ell,v,a\}$ &73.9 &   74.0   & 1.019 & 0.601 \\
BC-LSTM         & $\{\ell,v,a\}$ &75.2 &   75.3   & 1.079 & 0.614 \\ 
TFN             & $\{\ell,v,a\}$ &74.6 &   74.5   & 1.040 & 0.587   \\ 
GME-LSTM(A) 	& $\{\ell,v,a\}$ &76.5 & 73.4 & 0.955 & - \\
MARN 			& $\{\ell,v,a\}$ &77.1 & 77.0 & 0.968 & 0.625 \\ 
MFN				& $\{\ell,v,a\}$ &77.4 & 77.3 & 0.965 & 0.632 \\
LMF     		& $\{\ell,v,a\}$ &76.4 & 75.7 & 0.912 & 0.668 \\
RMFN     		& $\{\ell,v,a\}$ &78.4 & 78.0 & 0.922 & \textbf{0.681} \\
\Xhline{0.5\arrayrulewidth}
{\ours}      	& $\{\ell\}$ &\textbf{79.3}	& \textbf{79.1}	& \textbf{0.909}  & 0.676 \\ 
\Xhline{3\arrayrulewidth}
\end{tabular}
\caption{Sentiment prediction results on CMU-MOSI. Best results are highlighted in bold.
\ours \ outperforms the current state-of-the-art across most evaluation metrics and uses only the language modality during testing.}
\label{mosi_supp}
\vspace{-2mm}
\end{table}
\newcolumntype{K}[1]{>{\centering\arraybackslash}p{#1}}

\begin{table}[tb]
\fontsize{7.5}{10}\selectfont
\setlength\tabcolsep{1.0pt}
\begin{tabular}{l : c : *{2}{K{1.25cm}} : *{2}{K{1.25cm}}}
\Xhline{3\arrayrulewidth}
Dataset & & \multicolumn{2}{c:}{\textbf{ICT-MMMO}} & \multicolumn{2}{c}{\textbf{YouTube}} 
\\
Model        	& Test Inputs & Acc($\uparrow$) & F1($\uparrow$) & Acc($\uparrow$) & F1($\uparrow$) 
\\ 
\Xhline{0.5\arrayrulewidth}
RF				& $\{\ell,v,a\}$ & 70.0 & 69.8 & 33.3 & 32.3 
\\
SVM     		& $\{\ell,v,a\}$ &68.8 & 68.7 & 42.4 & 37.9 
\\
THMM			& $\{\ell,v,a\}$ &53.8	& 53.0 & 42.4 & 27.9 
\\
EF-HCRF		& $\{\ell,v,a\}$& 50.0 & 50.3 & 44.1 & 43.8 \\
EF-LDHCRF		& $\{\ell,v,a\}$& 73.8 & 73.1 & 45.8 & 45.0 \\
MV-HCRF		& $\{\ell,v,a\}$& 36.3 & 19.3 & 27.1 & 19.7 \\
MV-LDHCRF		& $\{\ell,v,a\}$& 68.8 & 67.1 & 44.1 & 44.0\\
CMV-HCRF		& $\{\ell,v,a\}$& 36.3 & 19.3 & 30.5 & 14.3  \\
CMV-LDHCRF		& $\{\ell,v,a\}$& 51.3 & 51.4 & 42.4 & 42.0 \\
EF-HSSHCRF		& $\{\ell,v,a\}$& 50.0 & 51.3 & 37.3 & 35.6 \\
MV-HSSHCRF		& $\{\ell,v,a\}$& 62.5 & 63.1 & 44.1 & 44.0 \\
DF   			& $\{\ell,v,a\}$ & 65.0	& 58.7 & 45.8 & 32.0 
\\
EF-LSTM   		& $\{\ell,v,a\}$& 66.3 & 65.0  &44.1 & 43.6 \\
EF-SLSTM		& $\{\ell,v,a\}$& 72.5	&70.9 & 40.7 & 41.2	 \\
EF-BLSTM 		& $\{\ell,v,a\}$& 63.8 & 49.6 & 42.4 & 38.1  \\
EF-SBLSTM 		& $\{\ell,v,a\}$& 62.5 & 49.0 &37.3 & 33.2 \\
MV-LSTM			& $\{\ell,v,a\}$ &72.5 & 72.3 & 45.8 & 43.3 
\\
BC-LSTM    		& $\{\ell,v,a\}$ &70.0 & 70.1 & 45.0 & 45.1 
\\ 
TFN      		& $\{\ell,v,a\}$ &72.5 & 72.6 & 45.0 & 41.0 
\\ 
MARN			& $\{\ell,v,a\}$ &71.3 & 70.2 & 48.3 & 44.9 
\\ 
MFN				& $\{\ell,v,a\}$ &73.8 & 73.1 & \textbf{51.7} & 51.6 
\\
\Xhline{0.5\arrayrulewidth}
{\ours} & $\{\ell\}$
& \textbf{81.3} & \textbf{80.8}
& \textbf{51.7}	& \textbf{52.4} 
\\ 
\Xhline{0.5\arrayrulewidth}
\Xhline{3\arrayrulewidth}
\end{tabular}
\caption{Sentiment prediction results on ICT-MMMO and YouTube. Best results are highlighted in bold.
\ours \ outperforms the current state-of-the-art across most evaluation metrics and uses only the language modality during testing.}
\label{full_supp}
\vspace{-2mm}
\end{table}

\end{document}